\documentclass[lettersize,journal]{IEEEtran}
\usepackage{amsmath,amsfonts}
\usepackage{algorithmic}
\usepackage{algorithm}
\usepackage{array}
\usepackage[caption=false,font=normalsize,labelfont=sf,textfont=sf]{subfig}
\usepackage{textcomp}
\usepackage{stfloats}
\usepackage{url}
\usepackage{verbatim}
\usepackage{graphicx}
\usepackage{multirow}
\usepackage{cite}
\usepackage{tikz}
\usepackage{gensymb}
\usepackage{xcolor}
\usepackage{hyperref}

\definecolor{lime}{HTML}{A6CE39}   
\DeclareRobustCommand{\orcidicon}{%
	\begin{tikzpicture}
	\draw[lime, fill=lime] (0,0) 
	circle [radius=0.16] 
	node[white] {{\fontfamily{qag}\selectfont \tiny ID}};
	\draw[white, fill=white] (-0.0625,0.095) 
	circle [radius=0.007];
	\end{tikzpicture}
	\hspace{-2mm}
}

\foreach \x in {A, ..., Z}{%
	\expandafter\xdef\csname orcid\x\endcsname{\noexpand\href{https://orcid.org/\csname orcidauthor\x\endcsname}{\noexpand\orcidicon}}
}

\begin{document}

\title{Micro-Fracture Detection in Photovoltaic Cells with Hardware-Constrained Devices and Computer Vision }

\author{Booy Vitas Faassen\orcidA{}, Jorge Serrano\orcidB{}, and Paul D. Rosero-Montalvo\orcidC{}

\thanks{Booy Vitas Faassen and Paul D. Rosero-Montalvo are with the Computer Science Department at the IT University of Copenhagen, Denmark. (email:
booy.f@outlook.com and paur@itu.dk).
Jorge Serrano is at OPTRONLAB, Department of Condensed Matter Physics, University of Valladolid, Spain (email: jorge.serrano@uva.es).}
}

\markboth{Journal of \LaTeX\ Class Files,~Vol.~14, No.~8, August~2021}%
{Shell \MakeLowercase{\textit{et al.}}: A Sample Article Using IEEEtran.cls for IEEE Journals}


\maketitle

\begin{abstract}
Solar energy is rapidly becoming a robust renewable energy source to conventional finite resources such as fossil fuels. It is harvested using interconnected photovoltaic panels typically built with crystalline silicon cells, i.e. semiconducting materials that convert effectively the solar radiation into electricity. However, crystalline silicon is fragile and vulnerable to cracking over time or in predictive maintenance tasks, which can lead to electric isolation of parts of the solar cell and even failure, thus affecting the panel performance and reducing electricity generation. This work aims to developing a system for detecting cell cracks in solar panels to anticipate and alert of a potential failure of the photovoltaic system by using computer vision techniques. Three scenarios are defined where these techniques will bring value. In scenario A, images are taken manually and the system detecting failures in the solar cells is not subject to any computational constraints. In this case, the InceptionV3 model, with a multilabel dataset, reached over 93\% accuracy. In scenario B, an Edge device is placed near the solar farm, able to make inferences. For these conditions, an EfficientNetB0 model shrunk into full integer quantization proved to be the most suitable model, reaching 85\% accuracy. Finally, in scenario C, a small microcontroller is placed in a drone flying over the solar farm and making inferences about the solar cells' states. In this situation, customized CNN architectures are the most suitable solutions since traditional ones are too big to be exported onto microcontrollers. Here, a machine learning model built with VGG16 blocks achieved 82\% accuracy on a binarized dataset.
\end{abstract}


\section{Introduction}
Climate change is a reality whose increasingly damaging effects call for action to governments, industry, and academia alike, embarking us on a quest for solutions to alleviate the effects and reduce there are other causes as well human-made root causes. The use of so-called green energy sources plays a fundamental role in this endeavour, by reducing the emission of pollutants into the atmosphere while maintaining economic growth. One of the most established, increasingly deployed solution technologies is the photovoltaic cell, which converts solar energy into a clean source of electricity~\cite{Gabor2006}. Nowadays, solar farms are widely deployed in large fields to power nearby cities. Additionally, smart homes, industrial facilities, and office buildings use solar panels to produce clean energy independently of the cities' electric system and store their produced energy in suitable batteries to consume afterwards~\cite{Stromer2019}. Specifically, solar energy has an investment of 20 billion dollars until 2030 around the world, and Germany and Spain are the European leaders in this energy production with 5.3 and 3.8 Gigawatts per year, respectively~\cite{Whitaker2020}. However, the performance of photovoltaic (PV) systems can be compromised by micro-fractures within the solar cells~\cite{Peng2014}. Micro-fractures are tiny cracks or defects in the crystalline structure of the cells that several factors, such as manufacturing imperfections, environmental stresses, transportation, installation, and long-term material degradation can cause. These damages can lead to localized areas of reduced efficiency that potentially spread over time, resulting in a significant decrease in the overall power output of the PV system~\cite{KajariSchroder2011}.

Detecting micro-fractures in PV cells timely is essential to ensuring the optimal performance and long-term reliability of solar energy generation. Traditional methods of micro-fracture detection involve manual and complex process inspections performed by trained personnel, which are time-consuming, labour-intensive, and prone to human error~\cite{Haase2018}. During in-field inspection, solar panels are sometimes removed from their place of installation to check these minor details in the solar cell, which could be a failure point when the solar panel is installed again~\cite{Hoffmann2020}. Conversely, specialized cameras and supporting equipment are used to take pictures without removing the solar panels and check their functionality~\cite{Spataru2017}. However, this procedure must be made in a controlled environment, often far from real operating conditions~\cite{KONTGES2011}.

To address the aforementioned challenges, computer vision methods can learn data patterns to infer these malfunctions in solar cells by using deep learning (DL) algorithms~\cite{Tsai2013}. To do this, it is necessary to have clean and annotated data, which requires experts to label the data and powerful servers to compile and train the DL models~\cite{Zheng2021}. Additionally, the external environment in which the DL models will operate must be defined with precision, since it provides the computational constraints that guide the selection of adequate hardware~\cite{sasaki2022}. Consequently, emerging technologies enable small devices such as microcontrollers to run inference close to the user or where the data is collected, providing energy savings and security benefits~\cite{bak2024}. However, finding a suitable model and training it from scratch is a time-consuming and inefficient process. Transfer learning is a technique where a convolutional neural network (CNN) pre-trained on a large image dataset can be reused in a new domain. To ensure the model fits the new domain, it is fine-tuned and re-trained on only a few sample images of this domain~\cite{Demosthenous2021}. In this paper, a dataset with images of solar cells with and without micro-fractures is used to re-train and fine-tune the machine learning (ML) models to detect micro-fractures~\cite{wu2020}. Subsequently, the models can be deployed on Edge devices to monitor and analyse the surfaces of solar cells in real-time. 

Designing a useful system, which includes choosing the right DL model and operating environment, is challenging due to hardware constraints~\cite{Zheng2021,Fagbohungbe2022}. 
This work explores three different environments, each subject to different hardware constraints, to find a suitable DL model. Additionally, we explore the feasibility of exporting pre-trained CNN models onto Edge or internet-of-the-things (IoT) devices. To this aim, we designed a novel machine learning methodology to accommodate the trade-offs between shrinking transfer-learned models and training small CNN models from scratch. The main contributions of this paper are as follows:
\begin{itemize}
  \item A comprehensive review of the existing methods and challenges of micro-fracture detection in PV cells is presented.
    \item An automated system that leverages emerging microcontrollers, Edge devices, and deep learning techniques for micro-fracture detection in real-time is proposed.
    \item The implementation details and key components of our approach, including data collection, image processing, and classification steps, are described.
    \item Finally, we report extensive experiments to evaluate the performance of our proposed system in terms of accuracy, efficiency, and robustness.
\end{itemize}

We defined three operating environments in which the ML models were deployed. In environment A, images were taken manually, and the system detecting micro-fractures is not subject to any computational constraints. As a result, the InceptionV3 model, with a multilabel dataset, reached over 93\% accuracy. Conversely, in environment B, where an Edge device was placed near the solar farm to make inferences on-site, an EfficientNetB0 model shrunk into full integer quantization proved to be the most suitable model by reaching 85\% accuracy. Finally, in environment C, where a small microcontroller was placed in a drone flying over the solar farm to make inferences about the solar cells' states, customized CNN architectures are the suitable solution since traditional ones are too big to be exported onto the hardware. Here, an ML model built with VGG16 blocks can achieve 82\% accuracy on a binarized dataset.

This article is structured as follows:
Section II surveys the related works. The machine learning pipeline is presented in Section III. The environmental setup and results are shown in Section IV. Finally, Section V is devoted to the conclusions and future work.

\section{Background and Related Works}
This section describes the solar panels' characteristics, classifies different types of damage, and summarizes other researchers' approaches to solar panel crack detection.

\subsection{Solar Modules}
Photovoltaic modules come in different sizes and consist of several PV cells. Different types of solar modules exist, with 95\% of modules sold consisting of crystalline silicon cells (c-Si) obtained from either single crystals (mono-Si) or polycrystalline wafers, referred to as poly-Si or multi-Si. In the latter, slightly lower efficiencies are obtained due to the effect of grain boundaries in carrier mean free path, i.e., 18\% vs 21\% efficiency in mono-Si. However, wafer production costs are reduced in poly-Si, yielding PV modules at a lower price~\cite{Energy.gov,Energysage}.

\subsection{Solar Cell Damages}
\begin{figure}
    \centering
    \includegraphics[scale = 0.75]{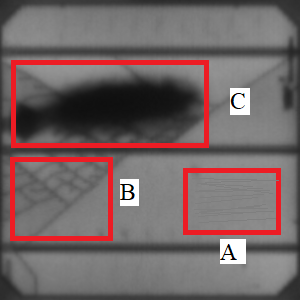}
    \caption{Electroluminiscence image of a single mono-Si solar cell containing type A, B, and C cracks.}
    \label{errors}
\end{figure}

Market demands to further reduce the cost of c-Si production drive manufacturers to reduce cell thickness, increasing the susceptibility to cracking and compromising the solar cells~\cite{KajariSchroder2011}. Cracks are formed as a mechanical load causes the module to bend, thereby jeopardizing the crystalline cells. If the pressure exceeds the cell strength, this will break from one edge to the other. Fractures can mechanically separate a part of the cell from the whole~\cite{marc2014}, potentially resulting in inactive cell areas~\cite{Haase2018}. Cracks are commonly classified into three categories, A, B, and C, depending on the level of severity and the effects on cell performance~\cite{Gabor2006,KONTGES2011}. Type A cracks leave cell areas electrically connected and are indistinguishable when inspected using electroluminescence (EL). In this case, there is no contrast in the EL image intensity in the crack region. In type B, however, the EL intensity is reduced in cell areas adjacent to the crack. Type C cracks represent the worst-case scenario: either the entire cell or significant adjacent areas have been rendered inactive, even affecting the electrical performance of both the cell and the module~\cite{Spataru2017}. Figure~\ref{errors} presents an EL image of a damaged crystalline cell, illustrating all three crack modes. Type C cracks appear black in EL images as separated cell regions do not emit any photons when the EL image is taken.

\subsection{Imaging and Computer Vision Techniques}

Micro-fractures in solar cells can only be observed with special imaging techniques, such as EL and photoluminescence (PL). However, researchers and practitioners appear to favour EL imaging techniques, used in state-of-the-art imaging systems~\cite{dos2017}. Electroluminescence imaging is a common and accurate method for observing PV cell damage and is particularly suitable for detecting micro-fractures in crystalline silicon cells~\cite{Benatto2021,Lockridge2016}, as it captures all three crack types in solar cells~\cite{Spataru2017}. As solar cells generate current when exposed to illumination, they also generate illumination when exposed to current. In EL imaging, an electrical current is passed through the cells to excite electrons into the conduction band, driving the cells into forward bias. This causes them to emit near-infrared (NIR) light corresponding to the band gap of silicon. In traditional methods, the electrical current is supplied by an external power source connected to the PV module. The infrared radiation emitted by the cells is subsequently captured with either a cooled charge-coupled device (CCD) with an infrared filter, or an InGaAs detector or camera, naturally tuned to give larger efficiency in the NIR region~\cite{Israil2013,Johnston2015, Benatto2021}. Recent EL and PL developments explore currently the potential of daylight inspection of solar panels in combination with a bias switching method~\cite{Guada2019}.

Traditional fracture detection methods include statistical, structural, and filter-based approaches~\cite{Rahman2020}. Statistical methods are used to separate the images into different regions, similar to histogram analysis. Structural approaches, instead, take into account spatial arrangements and textures in images~\cite{Chen2018}. Lastly, filter-based methods consider anomalies at specific points of interest. An example of this is Otsu's method. In~\cite{Peng2014}, Otsu's method is used to separate images into foreground and background. Then, Hough line detection is used to identify cracks in the EL images. Similarly, Stromer {\em et al.}~\cite{Stromer2019} apply the Vesselness algorithm, commonly used in medical image segmentation, to fracture segmentation, in particular, in polycrystalline cells. However, these methods are often inaccurate due to the vast amount of noise, particularly in polycrystalline cells~\cite{Rahman2020}. Tsai {\em et al.}~\cite{Tsai2013} propose using independent component analysis (ICA) for fracture detection, achieving a 93.4\% accuracy. This algorithm, however, classifies in a similar way other kinds of damages, such as finger interruptions, that do not cause a reduction in cell performance. In a pioneering study in 2019, Deitsch et al.~\cite{Deitsch_2019} utilize CNN models to automate solar cell defect detection. The authors compare two methods: feature detection in a support vector machine (SVM) and a deep CNN, achieving accuracies of 82.4\% and 88.4\%, respectively, on both poly- and monocrystalline cells. The CNN architecture is largely inspired by the Visual Geometry Group (VGG)-19 network, in which only the last layer on the top of the CNN architecture is customized to the specific domain. With weights pre-trained on ImageNet and fine-tuned on EL images, the SVM and CNN perform similarly on monocrystalline cells.

Xu {\em et al.}~\cite{xu2022celldefectnet} developed an attention condenser network specifically for EL-image fault detection on the Edge, called the CellDefectNet. The model architecture design was derived from a machine-driven exploration with generative synthesis~\cite{wong2018ferminets}. 

\section{Machine learning Pipeline}
\begin{figure*}
    \centering
    \includegraphics[scale=0.32]{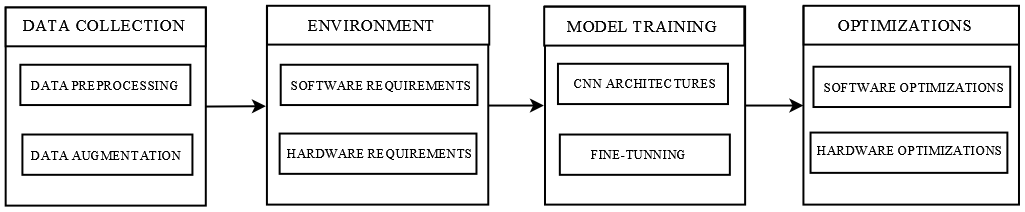}
    \caption{Novel ML pipeline to detect micro-fractures in solar cells.}
    \label{methodology}
\end{figure*}

The proposed methodology aims at developing
an ML model suitable for computationally constrained environments. It starts with the data collection and a pre-processing step. Then, we conceptually construct three environments to establish the hardware requirements. Several models are subsequently trained, and CNN architectures are defined. Their optimizations are deployed to run on small devices. Thereafter, the models are optimized to comply with the applicable hardware requirements. Finally, the ML models are evaluated on classification and performance metrics to shortlist the most suitable models. Figure~\ref{methodology} displays the proposed methodology.

\subsection{Data Collection}
\begin{figure*}[]
 \centering
  \subfloat[Mono 0.0]{
    \includegraphics[scale=0.19]{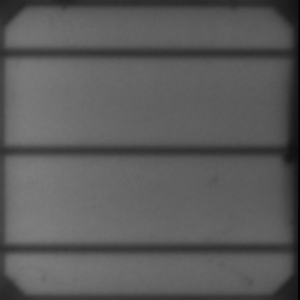}}
    \subfloat[Mono 0.3]{
     \includegraphics[scale=0.19]{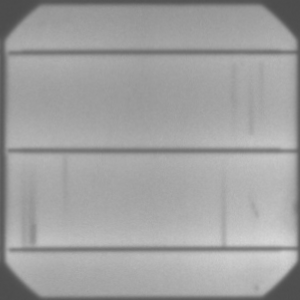}}
     \subfloat[Mono 0.6]{
     \includegraphics[scale=0.19]{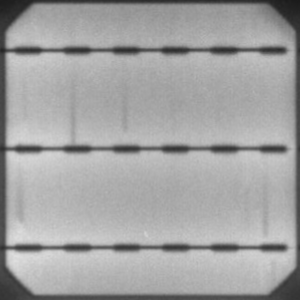}}
    \subfloat[Mono 1.0]{
     \includegraphics[scale=0.19]{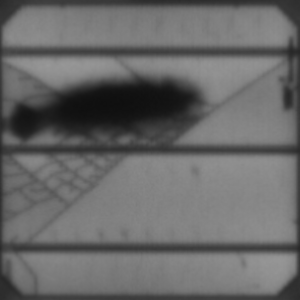}} 
    \subfloat[Poly 0.0]{
     \includegraphics[scale=0.19]{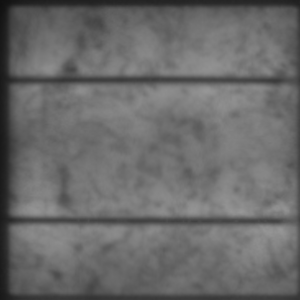}}
    \subfloat[Poly 0.3]{
     \includegraphics[scale=0.19]{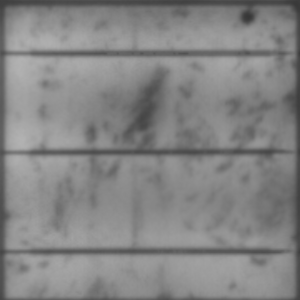}}
    \subfloat[Poly 0.6]{
     \includegraphics[scale=0.19]{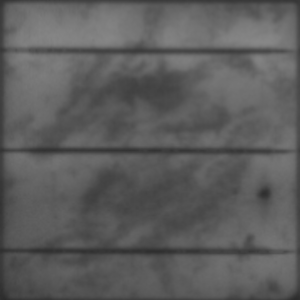}}
         \subfloat[Poly 1.0]{
     \includegraphics[scale=0.19]{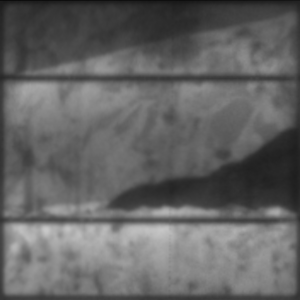}}
     \caption{Example images of each label. 0.0 means a fully functional solar cell, 1.0 is a wholly damaged solar cell, and  0.3 and 0.6 are not confident because the expert had doubts about annotating data.}
 \label{labels}
\end{figure*}

The dataset contains 2,624 samples of 300x300 pixels 8-bit grayscale images of functional and defective solar cells with varying levels of damage taken from 44 solar modules. Hereof, 26 were from polycrystalline modules, whereas 18 were from monocrystalline modules~\cite{Buerhop2018}. Several experts annotated the images and set a probability for that the solar panel contained damages. The defect label is set between 0 to 1, and also contains the type of solar module. As a result, the dataset has four labels for each solar module type; 0.0 means a fully functional cell, whereas 1.0 is a wholly damaged cell. 0.3 means likely functional but the expert was unsure, whereas 0.6 is likely damaged~\cite{Deitsch2019}. Figure~\ref{labels} shows images with different labels and solar module types.

For labels 0.3 and 0.6 the expert had doubts about annotating the data~\cite{Deitsch2021}. In addition, the dataset is unbalanced, and the labels mentioned above have a few samples. Therefore, we probed the label confidence by combining labels 0.3 and 0.6 with 0.0 to 1.0, respectively, to measure the classification accuracy and their impact on the classifier. Then, the dataset was converted to RGB images to use transfer learning techniques with pre-trained weights. Eight new datasets are produced and described in Table \ref{tab:datasets}. 

\begin{table}[]
    \centering
        \caption{Datasets descriptions related to several scenarios}
    \begin{tabular}{p{0.8 cm}|p{6 cm}|p {1 cm}} 
    \hline
       \textbf{Dataset}  & \textbf{Description} & \textbf{Total images}  \\
       \hline
         01 & Ratings 0.3 and 0.6 are assigned to 0.0 and 1.0, respectively. Stratified sampling of the ratings and types.& 2624  \\
         \hline
         02 & Ratings 0.3 and 0.6 are assigned to 0.0 and 1.0, respectively. Only images of type Mono-Si. & 1074 \\
         \hline
         03  & Ratings 0.3 and 0.6 are assigned to 0.0 and 1.0, respectively. Only images of type Poly-Si. & 1550 \\
         \hline
         04 & Ratings 0.0 and 1.0 are used for training, whereas ratings 0.3 and 0.6 are used for testing. Only images of type Mono-Si. & 1074 \\
         \hline
         05 & Ratings 0.0 and 1.0 are used for training, whereas ratings 0.3 and 0.6 are used for testing. Only images of type Poly-Si. & 1550 \\
         \hline
         06 & Ratings 0.3 and 0.6 have been removed. Images of both Mono-Si and Poly-Si. Stratified sampling of types and ratings. & 2223 \\
         \hline
         07 & Ratings 0.3 and 0.6 have been removed. Only images of type Mono-Si. & 9021 \\
         \hline
         08 & Ratings 0.3 and 0.6 have been removed. Only images of type Poly-Si. & 1322 \\
         \hline
    \end{tabular}
    \label{tab:datasets}
\end{table}
\subsection{Setting Environments}
\begin{figure}
    \centering
    \includegraphics[scale=0.35]{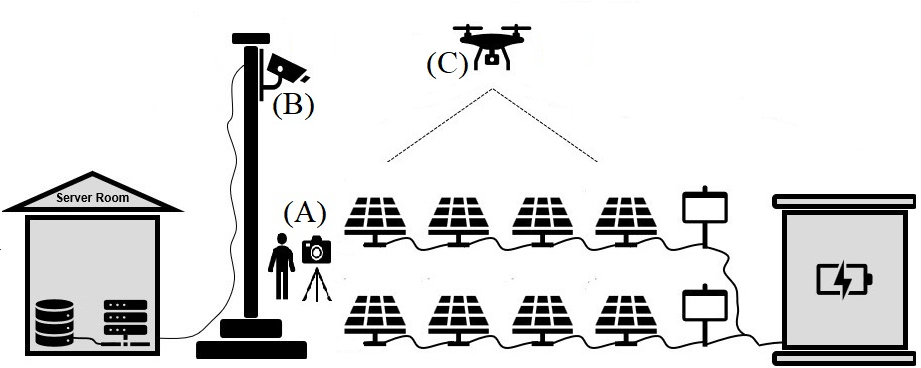}
    \caption{Environments' configurations to detect cracks in solar cells in a real-world setting.}
    \label{fig:environments}
\end{figure}

Given the many ways to detect cracks in solar cells on the field, we defined three different environments in accordance with tools, human support, and computational capabilities. Environment A is a system design in which a steadily placed camera takes images of solar cells. All images are manually uploaded to an external computer running the DL model. The camera, placed on a tripod, is temporarily and steadily stationed in front of the solar panel. At the same time, an external battery is connected to the panel and puts the cells in forward bias such that they emit light. This highly manual system resembles a quality control or inspection mechanism already present in solar farms today. In this environment, computations are not conducted on the Edge, and hardware does not place significant constraints on computing power, memory, or power consumption. 
In environment B, a stationary camera is permanently placed on a pole such that it overlooks the solar farm, whereas classification also occurs on the Edge. Since Edge devices are hardware-constrained, only the model's inferences could run on them. Therefore, the system detects failures in solar cells and communicates only the location of the cracked cell. To place the cells in forward bias, a camera-mounted laser [28] points at the cell and induces a voltage to spread, illuminating the cell. A Short-wave infrared (SWIR) camera can then capture the emitted irradiation~\cite{Israil2013, Alves2020}. The camera likely captures entire panels, and images must be pre-processed and cut into smaller images of the individual photovoltaic cells. Lastly, in environment C, a drone hovers over a solar farm, taking EL images of the solar panels and classifying them in real time. The system is highly autonomous and therefore enables continuous and automated inspection. The inference is run on the drone itself on a microcontroller, significantly constraining the system's memory and computational complexity. When a fractured cell is identified, only the coordinates of the cell are transmitted to an external server. Therefore, the transmission channel is significantly alleviated. The drone has the same camera configuration as Environment B. Figure \ref{fig:environments} shows the environments' configurations. 

\subsection{Model Training}

Different DL models were considered according to the set environments (A, B, and C) and their computational capabilities. Therefore, in environment A, large CNN models were trained due to their robustness and ability to predict correct labels in incoming data. In addition, these models were used to define the dataset that has the higher classification score to fit light DL models in the rest of the environments. In this scenario, we took new CNN algorithms that proved better than old CNN versions. Consequently, EfficientNets are constructed from the same building blocks, but their compositions and sizes differ. They were primarily built from the mobile inverted bottleneck residual blocks from MobileNetV2 and squeeze-and-excitation (SE) blocks~\cite{howard2017mobilenets}. EfficientNet-B0 comes at the lowest cost and complexity with 237 layers, whereas EfficientNet-B7 performs on the other side of the spectrum with 813 layers~\cite{tan2020efficientnet}. We selected EfficientNet B0 since the images from IoT datasets are small and this model weighs less than the others.
Conversely, Inception's architecture uses parallel convolutions that are subsequently concatenated to form a single module. The version of an Inception-V1 module consists of 1x1, 3x3, and 5x5 convolutions and a 3x3 max pooling layer in parallel. The feature maps of these layers are then concatenated before they are fed into the next module. The non-naïve version applies dimensionality reduction by adding 1x1 convolutions prior to the 3x3 and 5x5 convolutions after the pooling operation~\cite{szegedy2015rethinking,Feng2024}. To mitigate the vanishing gradient problem and provide additional regularization, two auxiliary classifiers are stacked on top of two Inception modules. These consist sequentially of an average pooling, a 1x1 convolution, a fully connected layer, a dropout layer, and Softmax activation~\cite{simonyan2015deep}. \\

Regarding environments B and C, light models, such as MobileNets, have a CNN with 53 convolutional layers. The architecture is based on an inverted residual structure with residual connections between the thin bottleneck layers. In addition, Squeezenet is a minor CNN architecture with a simple convolutional module design, using only 1x1 and 3x3 convolutional layers. Both models were designed specifically for resource-constrained environments and tested to run into ARM-based boards. Lastly, given that some microcontrollers can not run these models yet due to their RAM limitations,  we defined a final step to building tiny CNN architectures from scratch to run them into target microcontrollers. 

\subsection{Optimizations}

For the models trained for environments B and C, ARM-based boards must run inference of the DL model. Therefore, the models are optimized with four different quantization techniques from the TensorFlow Lite framework. To avoid an additional training overhead, given the resource constraints during training, only post-training quantization is used. Specifically, two techniques are applied: 16-bit floating-point quantization and 8-bit integer-quantization. Then, DL models are tested to check their abilities to accurately classify images (accuracy, precision, recall, F1-score) for the software side. On the hardware side, hardware utilization metrics, such as memory footprint, execution time, and power consumption were considered to select the most suitable DL model in each environment. 

\section{Environment setup and Results}
For the environment setup to run the ML models, we used a high-performance computing (HPC) cluster with a cluster with an Intel Core i7-4790 CPU powered with 8 cores, and 32 GB memory at 3.60~GHz. TensorFlow (TF) was selected as a framework to quantize models with TensorFlow-lite libraries. Datasets were divided into training, validation, and test sets using the cross-validation technique, thus splitting the dataset several times into the subsets mentioned above to get balanced results, avoiding bias or overfitting the model. 

\subsection{Environment A}

\begin{figure}
    \centering
    \includegraphics[trim={0.2cm 0.2cm 0.2cm 0.2cm},clip,scale=0.55]{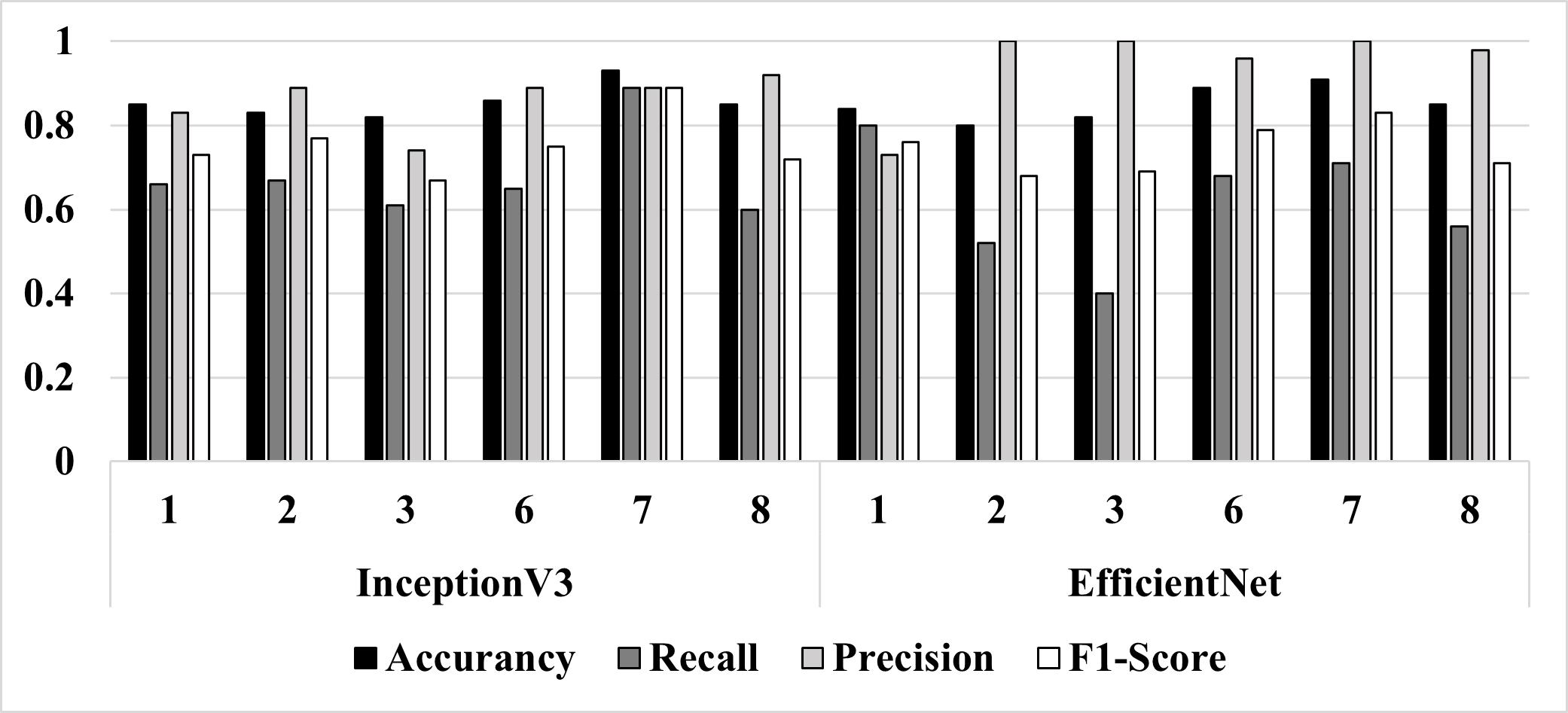}
    \caption{Classification accuracy of DL models set in environment A with second model configuration.}
    \label{fig:envA}
\end{figure}

In this environment, we used three model configurations. In the first configuration, models were added a Flatten Layer and Dense Layer to create the classifier on top of the CNN architectures. In the second configuration, ML models were fine-tuned by unfreezing some of the last layers of the feature extractor to fit the neuron's weight to the new domain, adding a Flatten Layer and Dense Layer on the top of the CNN architecture. In the third configuration, models had the same configuration as the previous one. However, we used data augmentation techniques to create synthetic data, such as rotation, translation, flip, and contrast. Figure~\ref{fig:envA} shows results in terms of model-centric metrics, such as accuracy, recall, precision and F1-score. In datasets 01 to 03, where the DL models were fed with multiclass data (labels 0.1, 0.3, 0.6 and 1.0), the model's accuracy is acceptable, mostly when the data is split in poly-Si and mono-Si types. Datasets 04 and 05, where the validation and test sets were only fed with labels 0.3 and 0.6, the DL models got low metrics because these models have a bias to predict wrong labels, therefore, suggesting that some images were mislabelled.

Conversely, Datasets 06 to 08 got promising results to shrink models, reaching high accuracy values, except for MobileNet models because they got low model-centric metrics values on each model configuration. As a result, InceptionV3 is the most robust model and was selected to run on this environment with multiclass or binary datasets, since it reached over 90\% accuracy. Lastly, model configuration 1 does not recognize patterns on datasets since they have a few layers to train. Model configuration 2 showed the highest model-centric metrics scores since they have several layers to fine-tune the model. Model configuration 3 showed that data augmentation techniques are unsuitable for this data type. Figure~\ref{fig:envA} shows the model-centric metrics with the second model configuration.

\subsection{Environment B}

\begin{figure}
    \centering
    \includegraphics[trim={0.2cm 0.2cm 0.2cm 0.2cm},clip,scale=0.55]{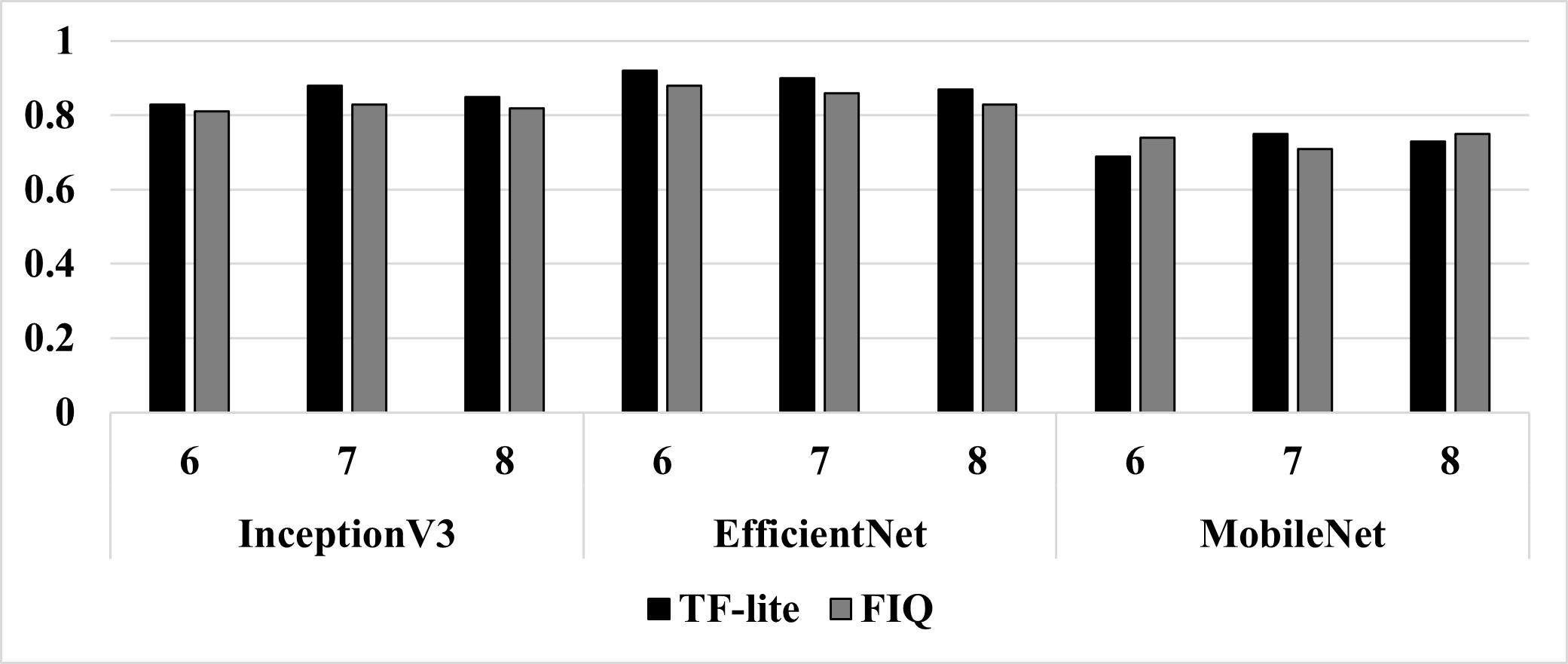}
    \caption{ TensorFlow-Lite (TF-lite) and full integer quantization performance (FIQ) analysis in environment B.}
    \label{envB}
\end{figure}

This environment, DL models were exported to make inferences on the device. TensorFlow-lite was the employed framework to shrink models with 32, 16 and full integer quantization resolution. However, shrinking the ML model with a 16-bit resolution did not show relevant advantages as the full integer quantization resolution does. Therefore, this technique was not taken into consideration in further sections. Table \ref{hardware} shows the target Edge devices with their relevant characteristics; some have small hardware accelerators to speed up the inference process. Since the DL model runs in a hardware-constrained environment working with a multilabel dataset, it is necessary to add Dense layers, increasing the model weight significantly and affecting the hardware-centric metrics. Therefore, we decided to move forward with a binary classification, which means DL models were trained with datasets 06 to 08. Figure~\ref{envB} shows how DL models perform using TF-lite (32-bit) and full integer quantization (8-bit) techniques to shrink the model. As a result, InceptionV3 and EfficientNet perform over 85\% on each dataset and MobileNet below 70\%. However, InceptionV3 is considered a heavy model since it weighs 91 Mbytes in the TF-lite version and 23 Mbytes in the quantized version. MobileNet is the lightest model, but it has the lowest classification accuracy, proving that it is not the right model to be deployed.

In contrast, EfficientNet weighs 16 Mbytes and 4.9 Mbytes for the TF-lite and full integer quantization, respectively, the latter having the highest classification accuracy. Consequently, this shrinked EfficientNet model was exported into the Edge devices to test their performance in real-trial settings since it had a similar classification performance to the TF-lite model with less memory footprint. As a result, the Jetson TX2 with its GPU can infer the label of new images in 5.56~ms, the Raspberry~Pi in 178~ms, the Jetson Nano in 13.58~ms, and the Coral dev could not run the model since it does not support some layers required by the model. 

  \begin{table}[ht!]
     \centering
          \caption{Edge devices hardware description}
     \begin{tabular}{p{1.5cm}|p{1.2cm}|p{1.2cm}|p{1.2cm}|p{1.2cm}}
     \hline
     & \multicolumn{4}{c}{\textbf{Edge devices}} \\
     \cline{2-5}
      \textbf{Specifications}  & Raspberry Pi 4  &  Jetson TX2 & Jetson Nano & Coral Dev     \\
    \hline 
    CPU &  Quad core Cortex-A72 64-bit SoC  & NVIDIA Denver, Quad-Core ARM Cortex-A57 & ARM Cortex-A57 MPCore & Quad Cortex-A53, Cortex-M4F \\
    \hline
    RAM & 4GB SDRAM & 8GB LPDDR4 & 4GB LPDDR4 & 1 GB LPDDR4 \\
        \hline
    Storage & Micro-SD card (64 GB) & 32GB eMMC &  Micro-SD card (64 GB) & Micro-SD card (64 GB)    \\
        \hline
    Camera Connector &  2-lane MIPI CSI camera port & CSI2 D-PHY 1.2 (2.5 Gbps/Lane) & MIPI-DSI x2 & MIPI-CSI2 camera input (4-lane)  \\
        \hline
    Hardware accelerator & None & NVIDIA Pascal GPU with 256 CUDA cores & 128-core GPU & Google Edge TPU:
4 TOPS (int8) \\
        \hline
     \end{tabular}
     \label{hardware}
 \end{table}

\subsection{Environment C}

Eventhough MobileNetV2 weighs 2.7 Mbytes when the model is quantized, this is a high computational requirement for microcontrollers, as they often have less than 1 Mbyte of RAM. SqueezeNet is a new small CNN architecture focused on microcontrollers. However, it needs larger computational resources to train the whole model than traditional desktops. In an exploratory analysis, this model did not reach 70\% of classification accuracy. Facing this scenario, we chose to build instead DL models from scratch, following both standard and new trends in CNN configurations. In addition, we set a requirement that the DL inference must weigh below 100 Kbytes to be exported in microcontrollers and leave enough RAM to take the image, call libraries and run the main program. Consequently, we used VGG16 and Inception blocks to build small CNN architectures using neural architecture research to get the suitable hyperparameter configuration and CNN architecture with as less memory footprint as possible. These DL models were trained with datasets 06 to 08 independently in order to get binary classification with datasets that contain either poly-Si or mono-Si cells. 

\begin{figure}
    \centering
    \includegraphics[trim={0.2cm 0.2cm 0.2cm 0.2cm},clip,scale=0.58]{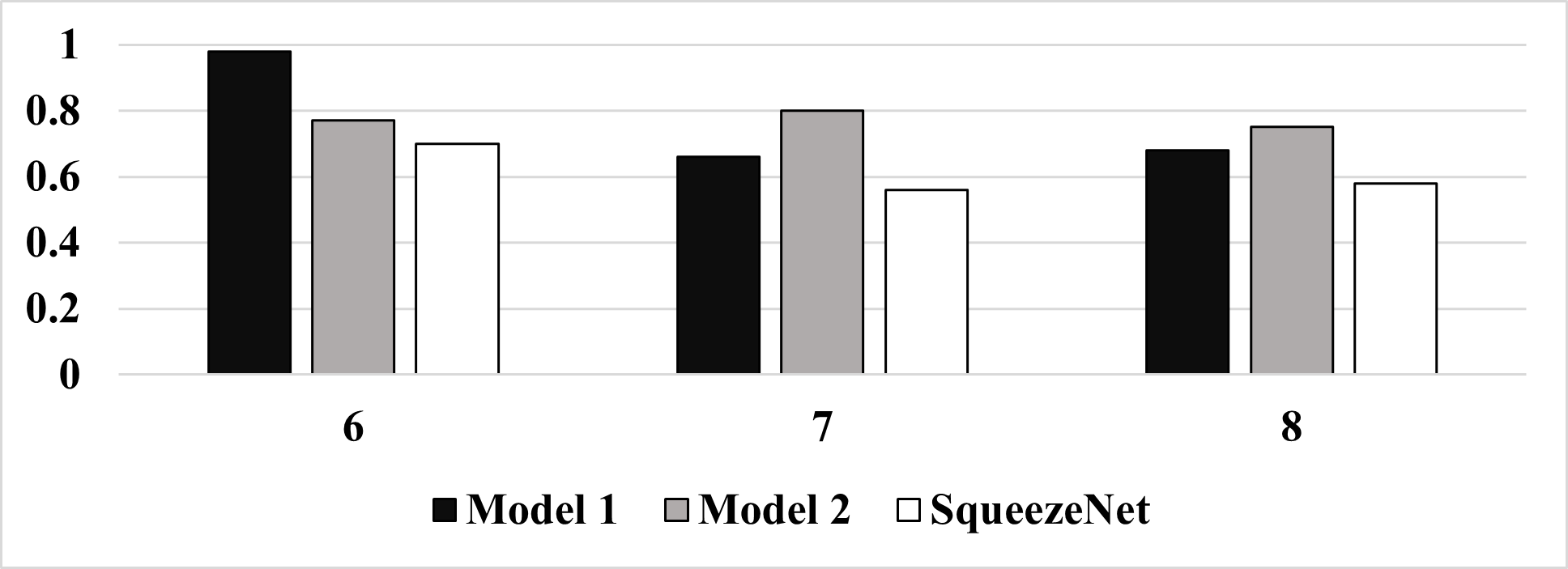}
    \caption{ Full integer quantization performance analysis in environment C.}
    \label{fig:envC}
\end{figure}

\begin{figure}[ht]
    \centering
    \includegraphics[scale=0.32]{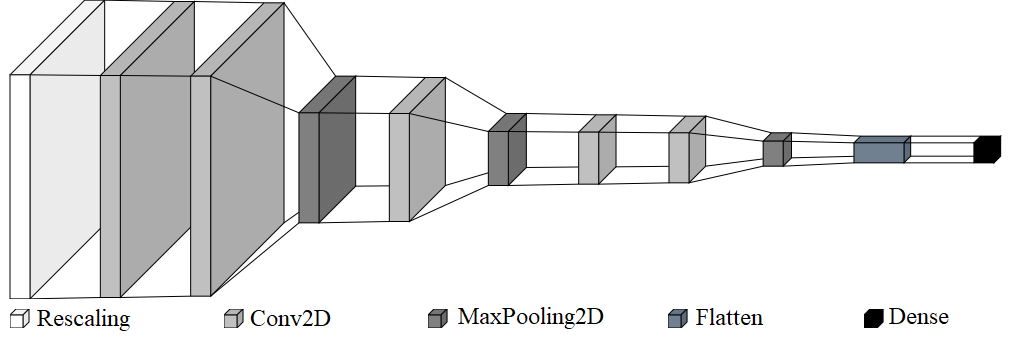}
    \caption{Model's architecture to detect cracks in PV cells with microcontrollers.}
    \label{fig:cmodels}
\end{figure}

Figure~\ref{fig:envC}. shows the model's performance. It reveals that model 1, built with the VGG16 blocks, performs over 82\% with dataset 06. However, it can detect a damaged or good solar cell without considering its type. Model 2 was built with Inception blocks. It performs below 80\% with the whole dataset. SqueezeNet performs poorly on each dataset. As a result, since this environment is focused on making quick alerts to take further actions, model 1 was selected to be deployed into the target microcontrollers. Figure~\ref{fig:cmodels}. shows the model architecture to reach solar crack detection with over 82\% accuracy. The target microcontrollers are described in Table \ref{hardware_uc}. \\

 \begin{table}[ht!]
     \centering
          \caption{Target microcontrollers' hardware description.}
     \begin{tabular}{p{1.8cm}|p{1.7cm}|p{1.7cm}|p{1.7cm}}
     \hline
     & \multicolumn{3}{c}{\textbf{IoT devices}} \\
     \cline{2-4}
      \textbf{Specifications}  & STM32H747XI  &  SAMD21  & ESP32-S   \\
    \hline 
    Microcontroller & Cortex®-M7 and Cortex®-M4  &  Cortex®-M0+ 32bit & ESP32-D0WD  \\
    \hline
    RAM (KB) & 1000  & 256  & 512  \\
        \hline
    Flash (MB) & 2 & 1 &  4  \\
        \hline
     \end{tabular}

     \label{hardware_uc}
 \end{table}

The Cortex-M7 can infer the label and process an image every 1.23~seconds, the Cortex-M0 in 2.34~seconds and the ESP32 in 2.44~seconds. Image acquisition and processing with the microcontroller took 600~ms on average because microcontrollers are slow compared to Edge devices. Therefore, since the Cortex M7 is a powerful microcontroller that can adequately handle the image pipeline, it is considered a suitable choice.

\section{Conclusions and Future works}
In the global transition to solar energy, effective and continuous inspection has largely remained unfeasible because of the environmental constraints in solar farms. This study presents several computer vision models to classify micro-fractures in photovoltaic cells. Specifically, various environmental constraints are accounted for, resulting in different model characteristics and capabilities. A 92.7\% accuracy is achieved using transfer learning to classify EL images of fractured PV cells in an unconstrained environment. Performance is improved by unfreezing layers,  modifying datasets, using different architectures, and retraining the classifier where applicable. Furthermore, weight quantization is necessary to reduce the model size and ensure hardware compatibility on microcontrollers and Edge devices. Quantization, where applicable, generally decreases accuracy, albeit at necessary memory requirements.

Moreover, transfer learning imposes limitations on model design that are particularly challenging for edge deployment. There is a trade-off between saving resources during training and creating a model suitable for Edge computing. Models deployed on the edge are already constrained, and transfer learning imposes additional limitations, particularly on the model architecture. The small MobileNetV2 architecture, quantized to 8-bit integers, is too large for deployment on an Arduino Portenta H7 microcontroller. Hence, alternative methods are required to reduce further memory footprint, such as developing custom CNNs. Edge deployment generally requires steps that reduce model performance, such as quantization and choosing smaller and shallower networks.

Furthermore, it is found that a traditional linear ML workflow is less suitable for deployment in severely constrained environments. Instead, the hardware requirements must be specified before selecting or building model architectures to avoid unnecessary re-work when optimized models are too large for a device. Given that small details in memory and performance matter, extensive experimentation is necessary. It is, therefore, essential to pre-select only those intuitively and potentially successful experiments rather than spending resources on architectures or methodologies that are eventually found unfeasible. 

Lastly, the study has several implications for academia and industry. First, the findings show solar plant operators that autonomous UAV-based inspection technologies are becoming a reality. However, small models generally come with reduced accuracy. Nonetheless, if only severely cracked cells must be identified, then the technology is soon readily available.

In future works, custom instructions to design a robust and tiny system-on-chip, SoC, with a small hardware accelerator, such as FPGAs, will be explored to deploy better CNN architectures in microcontrollers.

\section*{Acknowledgement}
J.S. acknowledges financial support from project grants PID2021-126046OB-C21/C22 funded by MCIN/AEI (Spain) and TED2021-130786B-I00 funded by MTED/AEI (Spain); both also funded by 10.13039/501100011033/FEDER, EU.

\bibliography{refs}
\bibliographystyle{IEEEtran}

\begin{IEEEbiography}[{\includegraphics[width=1in,height=1.25in] {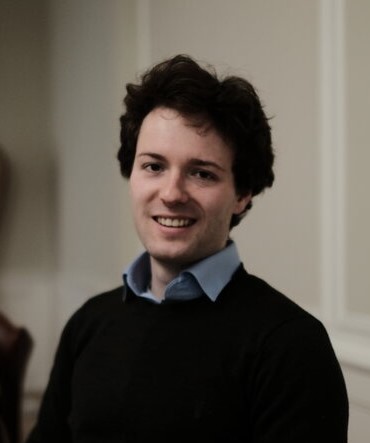}}]{Booy Vitas Faassen}
is a consultant specialised in formulating and executing on data and AI strategies in large corporations. He helps clients with utilising data and implementing AI into business processes. Booy has a Master of Science in Information Technology with a specialisation in Advanced Machine Learning from the IT University of Copenhagen (2023) and a Bachelor of Science in International Business from Copenhagen Business School (2021). For general inquiries contact: booy.f@outlook.com
\end{IEEEbiography}

\begin{IEEEbiography}[{\includegraphics[width=1in,height=1.25in] {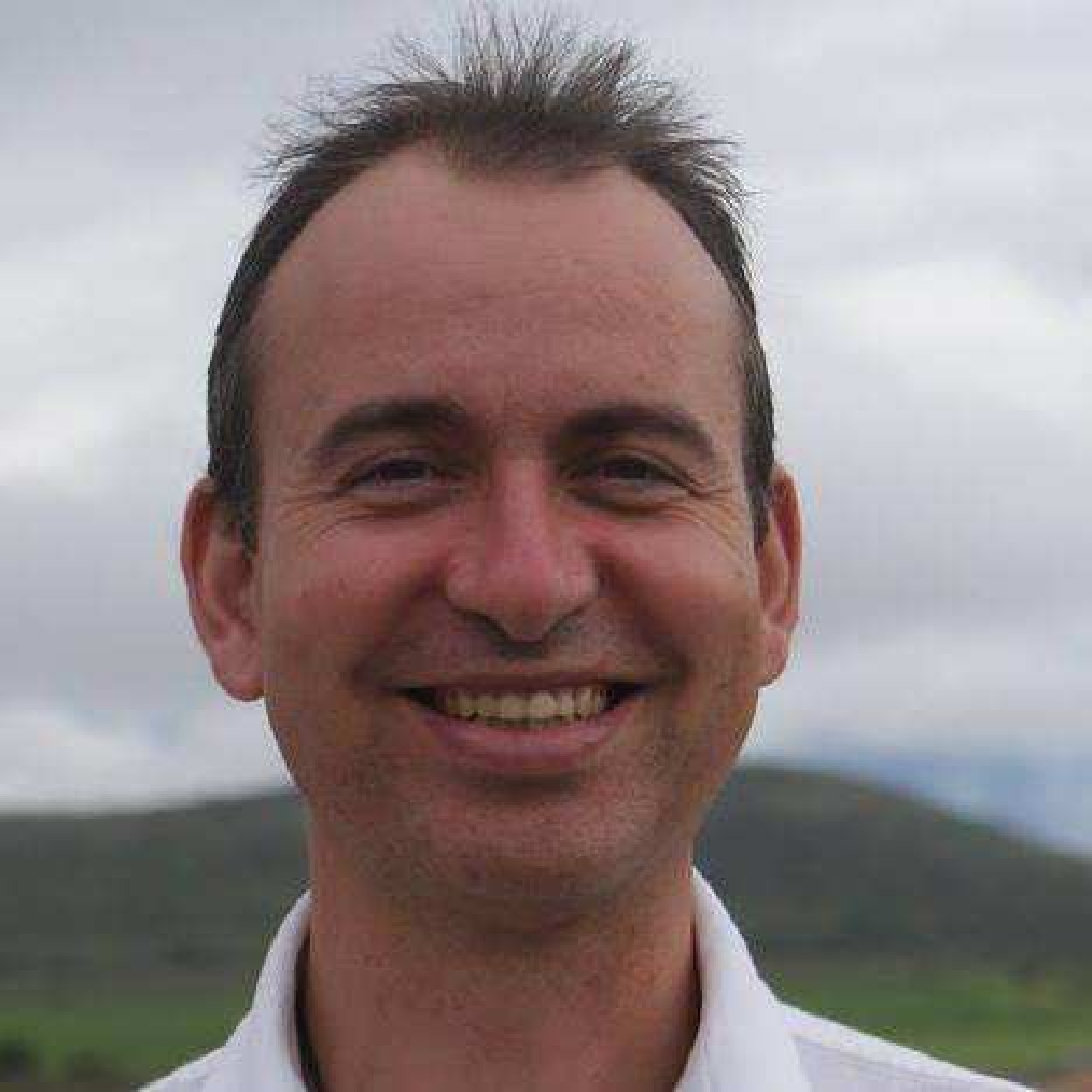}}]{Jorge Serrano}
He has a solid background in materials research and laser and x-ray spectroscopy, in parallel with  a background in education, professional coaching and leadership advising. He graduated in Physical Sciences at the University of Valladolid, Spain, and later was awarded a PhD at the University of Stuttgart, Germany, fruit of a productive research dedication at the Max Planck Institute for Solid State Research. He has conducted posdoctoral research at the École Polytechnique and the European Synchrotron Radiation Facility in France, and as ICREA researcher at the Universitat Polit{\'e}cnica de Catalunya in Spain, and at Yachay Tech University in Ecuador.  He is currently distinguished senior researcher at the University of Valladolid and combines his dedication to research with consultancy on leadership development in Science, Technology and Innovation, supporting brilliant leaders both in academia and in tech-based companies to excel even more in their career and expand a positive impact in the world.
\end{IEEEbiography}

\begin{IEEEbiography}[{\includegraphics[width=1in,height=1.25in] {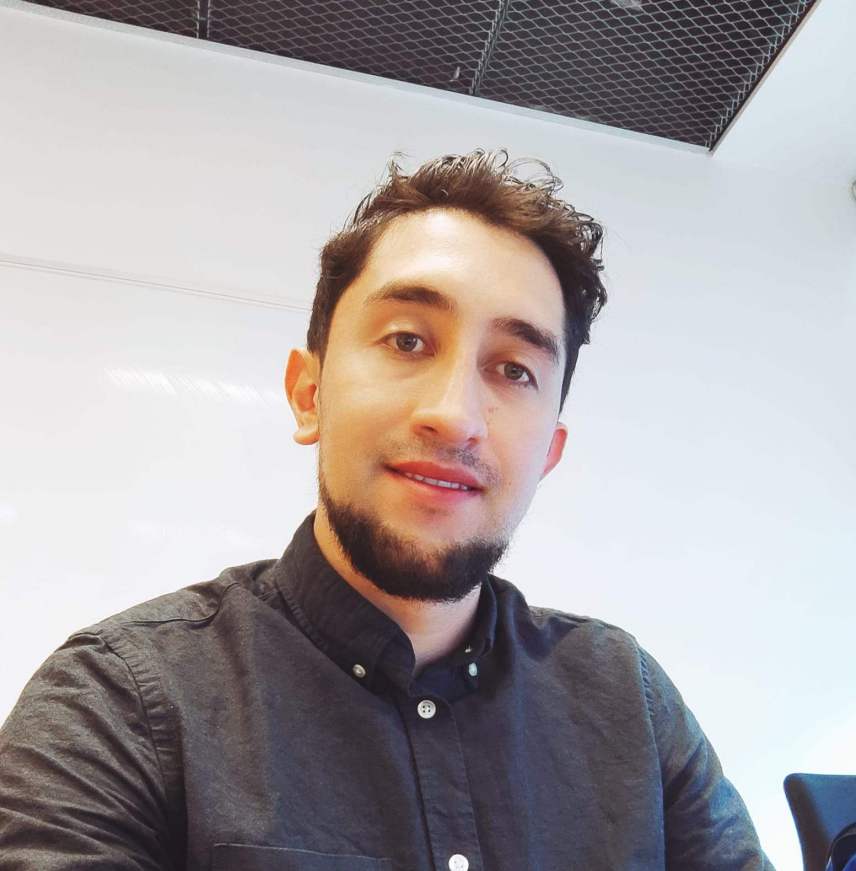}}]{Paul D. Rosero-Montalvo}

He is a Post. Doc at IT University of Copenhagen (ITU) in Denmark since June 2021. His research focuses on emerging microcontrollers to run machine learning models in decentralized networks. Earlier, he was a research assistant professor in the Applied Science Department at the Universidad Tecnica del Norte (UTN), Ecuador, for seven years. At the same time, he was a part-time lecturer in the TI Department at Instituto Tecnologico 17 de Julio, Ecuador. He received his Ph.D. from the University of Salamanca in Spain in November 2020, where Prof. Vivian Lopez-Batista advised him. He has a master's degree in Data Management systems from Universidad de las Fuerzas Armadas ESPE, Ecuador (2018), and an Engineering degree in Electronics and Telecommunications from UTN (2013).

\end{IEEEbiography}

\vfill

\end{document}